\documentclass[runningheads]{llncs}
\usepackage{graphicx}
\usepackage{hyperref} 
\usepackage{amsmath}
\usepackage{amssymb}

\usepackage{algorithm}
\usepackage{algorithmic}
\usepackage[table,xcdraw]{xcolor}
\usepackage{multirow}
\usepackage{graphicx}

\usepackage{bibentry}
\usepackage{verbatim}
\usepackage{subcaption}

\begin{document}
\title{Referring Industrial Anomaly Segmentation}

\author{Pengfei Yue\inst{1} \and
Xiaokang Jiang\inst{1} \and
Yilin Lu\inst{1} \and
Jianghang Lin\inst{1} \and
Shengchuan Zhang\inst{1}\thanks{Corresponding author.} \and
Liujuan Cao\inst{1}
}

\authorrunning{Pengfei Yue et al.}
%
\institute{
Key Laboratory of Multimedia Trusted Perception and Efficient Computing, Ministry of Education of China, Xiamen University, 361005, P.R. China.}

\maketitle              
%

\begin{abstract}
Industrial Anomaly Detection (IAD) aims to identify and locate anomalies in images, which is crucial for industrial manufacturing. 
Traditional unsupervised methods, which rely exclusively on normal data, produce simple classification results and rough anomaly localizations with manually defined thresholds. 
Meanwhile, supervised methods often overfit to prevalent anomaly types due to the scarcity and imbalance of anomaly samples.
Both paradigms suffer from the ``One Anomaly Class, One Model'' issue, which complicates practical applications. 
To tackle these issues, we propose Referring Industrial Anomaly Segmentation~(RIAS), a novel paradigm that leverages language to guide anomaly detection. 
RIAS offers two key benefits:
it generates precise, fine-grained masks directly from textual descriptions without the need for manual threshold adjustments, and it employs universal prompts to detect various anomaly types with a single, unified model.
To support RIAS, we introduce the MVTec-Ref dataset, designed to mimic real-world industrial environments with diverse, scenario-specific referring expressions.
This dataset is distinguished by two primary features. Firstly, the anomalies are unevenly distributed in terms of size, with small anomalies comprising 95\% of the dataset.
Secondly, in contrast to natural images that typically prioritize object recognition, the anomaly images in this dataset are designed to focus primarily on detecting the patterns of anomalies.
To evaluate the effectiveness of our paradigm and dataset, we propose a benchmark framework called \textbf{D}ual \textbf{Q}uery Token with Mask Group Transformer (DQFormer), which is enhanced by Language-Gated Multi-Level Aggregation (LMA). 
The LMA module enhances visual features at multiple scales, improving segmentation performance for anomalies of varying sizes. 
Additionally, unlike traditional query-based methods that rely on redundant queries, which are not well-suited for anomaly-focused images, our novel query interaction mechanism in DQFormer uses just two tokens, i.e. Anomaly and Background. This design facilitates efficient integration of visual and textual features.
Extensive experiments demonstrate the effectiveness of our RIAS in IAD.
We believe that RIAS, equipped with the MVTec-Ref dataset, will push IAD forward open-set. Code available at \href{https://github.com/swagger-coder/RIAS-MVTec-Ref}{https://github.com/swagger-coder/RIAS-MVTec-Ref}.

%

\end{abstract}

\section{Introduction}
\label{sec:introduction}
\begin{figure}[]
\centering 
\includegraphics[width=0.6\linewidth]{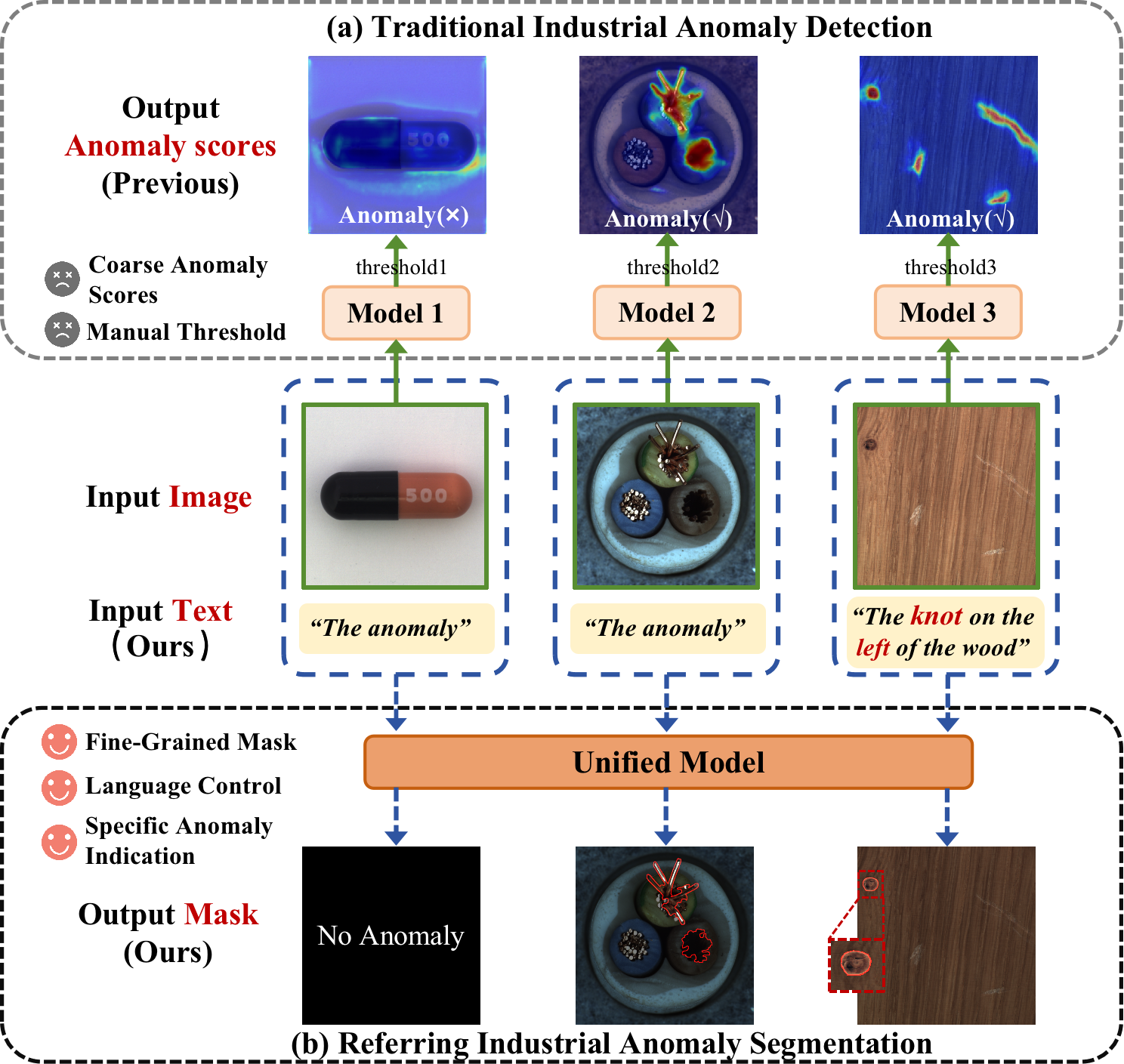} 
\hfill 
\centering
\caption{\textbf{Comparison between traditional IAD and our RIAS.} Traditional IAD methods generate coarse anomaly scores for localization, which also requires manual threshold adjustments to obtain the final classification result. Additionally, these methods suffer from the ``One Anomaly Class, One Model" issue. In contrast, RIAS produces precise, fine-grained masks based on flexible textual descriptions, eliminating the need for manual thresholds. RIAS also supports the use of universal prompts to detect any anomaly types within one single unified model.} 
\vspace{-15px}
\label{fig: motivation} 
\end{figure}

Industrial Anomaly Detection~(IAD) task aims to identify abnormal images and locate abnormal subregions. 
The technique to detect the various anomalies of interest has a broad set of applications in manufacturing applications, such as surface anomaly detection and textile defect detection~\cite{wang2022hybrid,bergmann2019mvtec}. 
Most existing methods for IAD are unsupervised, assuming that only normal~(anomaly-free) training data are available due to the practical challenges of collecting large-scale anomaly data~\cite{liu2023simplenet,zavrtanik2021draem}. 
While these unsupervised methods excel in classification, they often yield only rough anomaly localization by anomaly score, as depicted in Fig.\,\ref{fig: motivation}, lacking the precision for pixel-level detail.
Conversely, recently supervised methods utilize a limited set of labeled anomaly samples to train more informed models~\cite{ding2022catching,pang2021explainable}. 
These methods either employ one-class metric learning with anomalies as negative samples~\cite{liu2019margin,ruff2019deep} or utilize one-sided anomaly-focused deviation losses~\cite{pang2019deep,pang2021explainable}. However, their reliance on seen anomalies can lead to overfitting and poor generalization to novel anomaly types.
Furthermore, both unsupervised and supervised IAD methods typically necessitate a dedicated model for each anomaly category, which can be resource-intensive~\cite{shi2016uniad}. 
Given the inherently unpredictable nature of anomalies, IAD is fundamentally an open-set problem~\cite{ding2022catching}. To improve both the understanding and detection of anomalies, we propose leveraging linguistic cues.
We introduce Referring Industrial Anomaly Segmentation~(RIAS), a novel paradigm that integrates image and language to specify anomalies of interest. 
RIAS enables the identification of anomalies and the generation of pixel-level masks driven by language descriptions.

As shown in Fig.\,\ref{fig: motivation}, RIAS allows for the flexible generation of fine-grained masks, thereby enabling precise anomaly localization. For instance, 
``The knot on the left of the wood" specifies only a subset of anomaly types, requiring the model to distinguish between the knot and scratch categories within a combination of anomalies. 
As a result, our RIAS offers a new perspective on studying anomaly patterns.
Unlike the common setting of Referring Image Segmentation~(RIS), which assumes that the text corresponds to only one object~\cite{yue2024adaptive}. 
Following the setting of Generalized Referring Image Segmentation~\cite{liu2023gres}, RIAS further supports multi-target expression that specifies multiple target anomalies in a single expression, e.g. "The anomal". 
The universal prompts provides much more flexibility for input expression, giving RIAS greater generalization ability for novel anomaly types. Furthermore, this enables the use of one singe unified model to handle any anomaly types, effectively addressing the "One Anomaly Class, One Model" limitation in previous methods.
Additionally, we are the first to study the IAD domain from an Image-Language perspective, bringing significant research insights to both Image-Language understanding and Image anomaly detection fields.
To facilitate our groundbreaking research on RIAS, we introduce the MvTec-Ref dataset which is built upon the images and pixel-wise annotations from the MvTec-AD dataset~\cite{bergmann2019mvtec}. 
The MvTec-Ref dataset comprises $2,110$ image-language-label triplets. These triplets are carefully designed to capture various referring expressions that reflect actual anomaly scenarios, including categories, attributes, and spatial relationships within objects.
The dataset is distinguished by two primary features. Firstly, the anomalies are unevenly distributed in terms of size, with small anomalies comprising $90\%$ of the dataset.
Secondly, as illustrated in Fig.\,\ref{fig: motivation}, the anomaly images in this dataset inherently focus primarily on detecting the patterns of anomalies, which contrasts with natural images that typically prioritize object recognition.
To assess the efficacy of our RIAS, we propose a benchmark framework called \textbf{D}ual \textbf{Q}uery Token with Mask Group Transformer~(DQFormer), which is enhanced by Language-Gated Multi-Level Aggregation~(LMA).
The LMA module enhances visual features across multiple scales, thereby improving segmentation performance for anomalies of varying sizes. 
%
%
Furthermore, unlike traditional query-based methods that rely on redundant queries\cite{tang2023contrastive,liu2023gres,yang2022lavt}, which are not well-suited for anomaly-focused images, our novel query interaction mechanism in DQFormer uses just two tokens, i.e. Anomaly and Background. This design facilitates efficient integration of visual and textual features.
To comprehensively explore RIAS, we compare our proposed DQFormer with various RES methods on the MVTec-Ref dataset.
Our findings reveal that the DQFormer outperforms the best previous methods~\cite{tang2023contrastive} by $3.08\%$ in mIoU and $3.42\%$ in gIoU on the validation set, and by $2.5\%$ in mIoU and $2.16\%$ in gIoU on the test set, demonstrating its superiority. 
Additionally, we compared DQFormer with traditional IAD methods and found that DQFormer achieves the best performance within a unified model.
Extensive experiments further highlight the flexibility and effectiveness of our RIAS in IAD.



\section{Related Work}
\subsection{Industrial Anomaly Detection}
Existing IAD methods can be categorized into unsupervised and supervised approaches. Many existing unsupervised anomaly detection methods, including those based on autoencoders~\cite{gong2019memorizing}, GANs~\cite{perera2019ocgan}, self-supervised learning~\cite{bergman2020classification}, and one-class classification~\cite{chalapathy2018anomaly}, operate under the assumption that only normal data is available during training. While this approach avoids the risk of bias towards known anomalies, it often struggles to effectively distinguish anomalies from normal samples due to the lack of information about true anomalies. In contrast, there has been a recent shift towards supervised (or semi-supervised) anomaly detection, which addresses the lack of anomaly information by using a small number of labeled anomaly examples to train models that are informed by these anomalies. This is typically done through one-class metric learning~\cite{gornitz2013toward}, where anomalies are treated as negative samples, or by employing anomaly-focused deviation loss~\cite{pang2019deep,pang2021explainable}. However, these models are often prone to overfitting to the specific anomalies they have seen. 

\subsection{Referring Image Segmentation}
Traditional unimodal detection~\cite{lin2024weakly,mi2022active} and segmentation~\cite{lin2025you,lin2025pseudo} are insufficient for human interaction with the real world; thus, language-guided segmentation paradigms have gained increasing attention. Referring Image Segmentation~(RIS) is designed to localize objects within images guided by natural language descriptions. The early approach utilized a fusion technique combining linguistic and visual elements through concatenation~\cite{hu2016segmentation}. Subsequent efforts have harnessed sentence-level textual features from the descriptive phrases~\cite{li2018referring,chen2019referring}, whereas other studies have adopted word-level textual features for textual representation~\cite{li2018referring,chen2019referring}. Given that natural language intrinsically contains structured data that can be exploited to align with visual features~\cite{shi2022spatial}, certain methodologies have decomposed expressions into various components~\cite{hui2020linguistic} or implemented a soft division approach using attention mechanisms~\cite{ding2021vision,yu2018mattnet}. Recent work has adopted more efficient structures for vision-language fusion. LAVT~\cite{yang2022lavt} utilizes the Swin Transformer for visual tasks and incorporates modules for vision-language integration in the last four layers of the visual encoder. In contrast, ReSTR~\cite{kim2022restr} and CRIS~\cite{wang2022cris} start by separately encoding visual and linguistic inputs with a dual encoder, then merging these features either through a multi-modal transformer encoder or a cross-modal decoder. 
\section{Task Setting and Dataset}
\subsection{Task Setting}
Referring Industrial Anomaly Segmentation~(RIAS) is a novel task designed to leverage natural language descriptions for the segmentation of anomalies in industrial images. 
During training, the input sample consists of an image \( I \), a language description \( L \), a ground-truth segmentation mask \( M_{GT} \), and a no-anomaly identifier \( E_{GT} \). The mask \( M_{GT} \) identifies the correct regions of interest for the anomalies described in \( L \), while \( E_{GT} \) indicates whether the description \( L \) refers to any anomalies in the image. The model is tasked with outputting a corresponding mask \( M \) and a no-anomaly identifier \( E \) based on the text description.
During inference, if the no-anomaly identifier \( E \) is predicted to indicate no anomalies, the model outputs an empty mask.
%
%
\subsection{MVTec-Ref Dataset}
To facilitate our groundbreaking research on RIAS, we develop the MVTec-Ref dataset, an extension of the widely utilized MVTec-AD dataset for anomaly detection~\cite{bergmann2019mvtec}. 
The inherent complexity and diversity of anomalies often preclude effective classification into single categories. 
Consequently, we have standardized the annotation process to align more closely with the intrinsic characteristics of the anomaly data. 
The final Mvtec-Ref dataset consists of 2,110 image-language-label triplets. The overall annotation process, as well as data analysis of the dataset, is detailed further in the \textit{Appendix}.
\begin{figure}[]
\vspace{-10px}
\centering 
\includegraphics[width=0.7\linewidth]{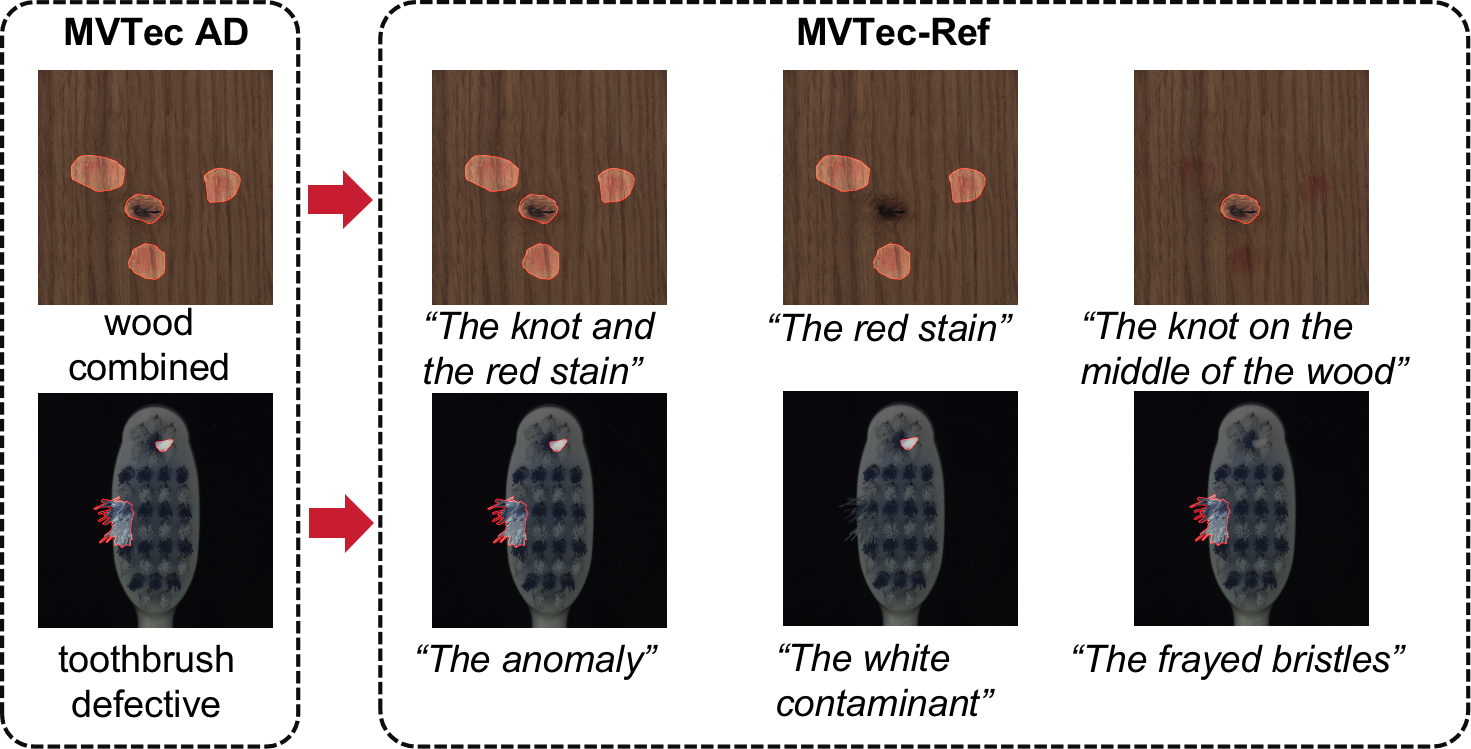} 
\hfill 
\centering
\caption{Examples from MvTec AD to MVTec-Ref.} 
\vspace{-50px}
\label{fig: dataset analysis} 
\end{figure}
\\
\\
\begin{figure*}[] 
\centering 
\includegraphics[width=\textwidth]{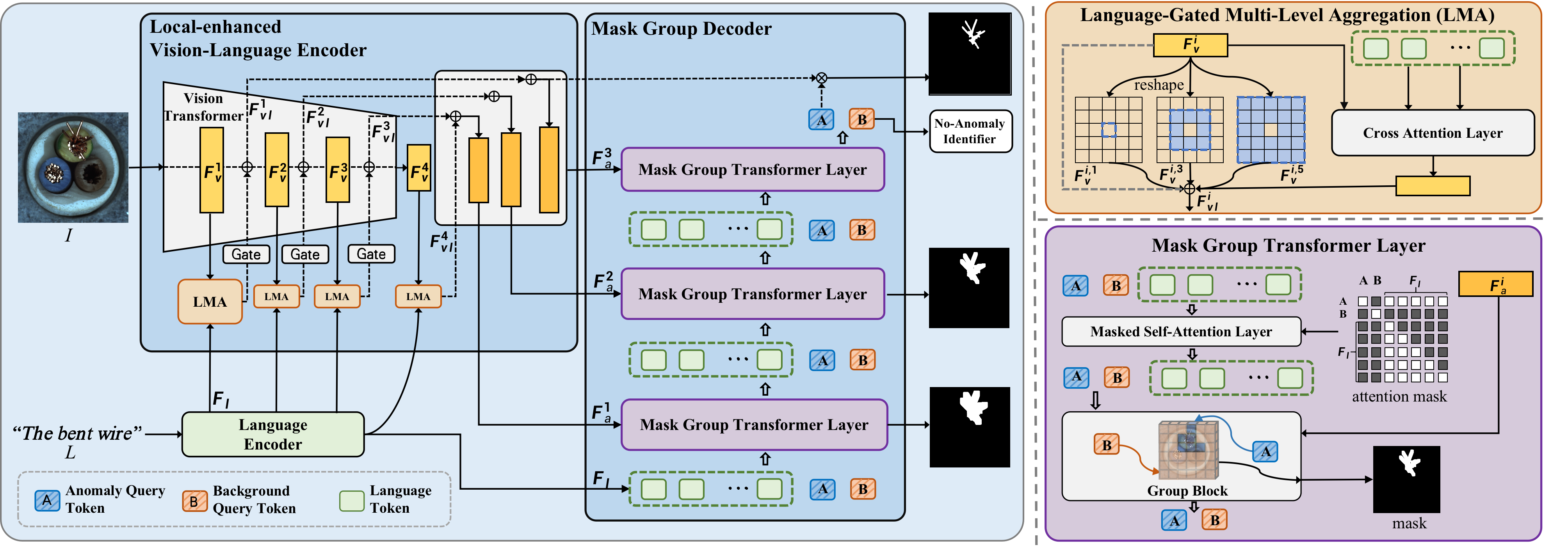} 
\caption{\normalfont The framework of DQFormer includes the overall framework, a Language-Gated Multi-Level Aggregation~(LMA) module, and a Mask Group Transformer Layer. Initially, the Local-Enhanced Vision-Language Encoder~(LVLE) takes \( F^i_v \) and \( F_l \) as inputs to the LMA, generating local-aware and language-aware features \( F^i_{vl} \). These \( F^i_{vl} \) features are then fused into the Vision Transformer through a gating mechanism. After a simple aggregation, they are fed into the Mask Group Decoder, where the Dual Query tokens \( A \) and \( B \) interact with language features $F_l$ and aggregated visual features $F^i_a$ sequentially.  Within each Mask Group Transformer Layer, only \( A \) interacts with the language features through a Masked Self-Attention mechanism. Finally, \( A \) is multiplied by \( F^3_a \) to generate the final mask output \( M \), while \( B \) is mapped to a No-Anomaly Identifier \( E \).
}
\vspace{-10px}
\label{fig: main} 
\end{figure*}
\section{Method}
\subsection{Overview}
An overview of our approach is illustrated in Fig.\,\ref{fig: main}. 
Initially, the expression $L$ is encoded to extract language features \( F_l \).
%
The Local-Enhanced Vision-Language Encoder~(LVLE) performs multi-level local enhancement on the visual features \( F_{v} = \left\{ F^i_{v} \right\}_{i=1}^{4} \) from the four stages \( i \in \{1,2,3,4\} \), then integrates the visual features \( F_{v}\) with the language features \( F_l \) to obtain local-aware and language-aware features \( F_{vl} = \left\{ F^i_{vl} \right\}_{i=1}^{4} \). Next, the LVLE aggregates the adjacent \( F_{vl} \) features across layers to produce a multi-scale aggregated feature \( F_{a} = \left\{ F^k_{a} \right\}_{k=1}^{3} \)~(see Section \ref{subsection: LVLE}).
Subsequently, the Dual Query tokens interact with both the language features \( F_l \) and the aggregated visual features \( F_{a} = \left\{ F^k_{a} \right\}_{k=1}^{3} \) within the Mask Group Decoder sequentially~(see Section \ref{subsection: Mask Group Decoder}).
%
Finally, the updated Anomaly Query Token \( A \), together with the highest resolution feature map \( F^3_a \), is then used to generate the mask output \( M \). Meanwhile, the Background Query Token \( B \) is mapped to predict a No-Anomaly Identifier \( E \in \mathbb{R}^{2} \), indicating whether an anomaly is present.
\subsection{Local-enhanced Vision-Language Encoder}
\label{subsection: LVLE}
The Local-enhanced Vision-Language Encoder~(LVLE) module mainly consists of the Vision Transformer and the Language-Gated Multi-Level Aggregation~(LMA) module. 
The Vision Transformer first extracts the visual features \( F_{v} = \left\{ F^i_{v} \right\}_{i=1}^{4} \) at four stages \( i \in \{1,2,3,4\} \).
Then, the Language-Gated Multi-Level Aggregation~(LMA) module first enhances the visual features with multi-level local feature aggregation. 
Afterward, it integrates these enhanced visual features with the language features \( F_l \) to obtain the multi-modal features \( F_{vl} = \left\{ F^i_{vl} \right\}_{i=1}^{4} \), which are then fed back into the Vision Transformer.
Next, the LVLE aggregates the adjacent \( F_{vl} \) features across layers to produce a multi-scale aggregated visual feature \( F_{a} = \left\{ F^k_{a} \right\}_{k=1}^{3} \), following the approach used in CGFormer~\cite{tang2023contrastive}.
%
\subsubsection{Vision Transformer} Following the previous work~\cite{yang2022lavt} , we employ Swin Transformer~\cite{liu2021swin} as our Visual Transformer. For an input image $I \in \mathbb{R}^{H \times W \times 3}$ with the size of $H \times W$, we extract its visual feature maps at four stage $i \in \{1, 2, 3, 4\}$, which we denote it as $F_v = \{F^{i}_v\}_{i=1}^4, F^{i}_v \in \mathbb{R}^{H_i \times W_i \times C_i^v}$. Here, each stage is corresponding to an encoding block of Swin Transformer, and $H_i$, $W_i$ and $C_i^v$ denote the height, width and channel dimension of $F^{i}_v$.

\subsubsection{Language-Gated Multi-Level Aggregation~(LMA)}
As shown in the upper-right corner of Fig. \ref{fig: main}, the LMA module has two branches. 
To improve the modeling of anomalies in objects of varying sizes, the left branch performs multi-level local feature aggregation on the visual features \( F^i_v \) extracted from the \( i \)-th block of the Vision Transformer. 
In this branch, as depicted in the LMA module in Fig. \ref{fig: main}, an adaptive average pooling operation is applied within an \( r \times r \) neighborhood, where \( r \) is selected from \{1, 3, 5\} to obtain the multi-level enhanced features.
The average pooling operation enables us to achieve hierarchical perception of neighboring features. For the feature at position \((x,y)\) in \( F^i_v \), this process can be represented by the following equation:

\begin{equation}
F^{i,r}_v[x,y] = \frac{1}{r \times r} \sum_{m=0}^{r-1} \sum_{n=0}^{r-1} F^i_v[x \cdot s + m, y \cdot s + n],
\end{equation}
where \(F^i_v[x, y]\) denotes the value of the input feature map at position \((x, y)\). The term \(r \times r\) indicates the size of the pooling window, while \(s\) denotes the stride, which is set to 1 in our implementation. We achieve the aggregation of multi-level local features by summing element-wise: $F^i_{t} = \sum_{r \in \{1, 3, 5\}} F^{i,r}_v,$
where \(F^{i, r}_{t} \in \mathbb{R}^{H_i \times W_i \times C_i^t} \) represents the local-enhanced visual features with level $r$. 

In the right branch, to model global visual-linguistic relationships, we employ a gated cross-modal attention mechanism following LAVT~\cite{yang2022lavt}. This involves projecting and fusing the visual features \(F^i_v\) with the language features \( F_l \) through cross-modal attention~\cite{vaswani2017attention}. The final fused features \( F^i_{vl} \in \mathbb{R}^{H_i \times W_i \times C_i^t} \) are obtained through the following operation:
\begin{equation}
\begin{aligned}
 F^i_{vl} &= \text{CA}(F^i_v, F_l) + F^i_t,
\end{aligned}
\end{equation}
where \( \text{CA} \) is the cross-attention operation, $F^i_t $ is the local-enhanced visual features of $i$-th block. 
Once $F^i_{vl}$ is obtained, similar to \cite{yang2022lavt}, we combine the $F_{vl}$ from LMA with visual features $F^i_v$ from the Transformer layers. This process employs a gating mechanism, $S_i$, which learns a set of element-wise weight maps from $F_{vl}$ to finely tune the scale of each element within $F^i_{vl}$. Subsequent to this adjustment, a residual combination of $F^i_v$ and $F^i_{vl} \cdot S_i$ is performed. The resultant output is then fed back into the Visual Transformer for further computation.


We then propose an efficient multi-scale strategy to utilize high-resolution features~\cite{tang2023contrastive}. After obtaining the multi-scale visual-language features \( F^i_{vl} \) from the \( i \)-th stage, we employ a lightweight multi-scale feature aggregator. Specifically, we merge features from adjacent scales:
\begin{equation}
\begin{aligned}
    F^k_a &= \text{Conv}([F^k_{vl}; \text{Up}(F^{k-1}_a)]),  k \in \left\{2, 3, 4 \right\},
\end{aligned}
\end{equation}
where $F^1_a = F^1_{vl}$, $\text{Up}(\cdot)$ denotes the upsampling operation that aligns the lower resolution feature map $F^{k-1}_{a}$ to the higher resolution feature map $F^k_{vl}$, $[;]$ indicates that the two feature maps are concatenated along the channel dimension.
Then $F^{i}_{a}$ is input into Mask Group Transformer layer by layer according to the resolution size.
\subsection{Mask Group Decoder}
\label{subsection: Mask Group Decoder}
The Mask Group Decoder employs a three-layer Mask Group Transformer to progressively refine mask outputs. Initially, the Dual Query tokens, i.e. Anomaly Query Token \( A \) and Background Query Token \( B \), are fed into the Mask Group Decoder.
In the \(k\)-th layer of the Mask Group Transformer, these tokens first interact with the language features \( F_l \), with a design that ensures only the Anomaly Query Token \( A \) interacts with the language features \( F_l \).
Subsequently, the Dual Query tokens engage with the aggregated visual features \( F^k_{a} \) within the Group Block, which is designed following the principles of CGFormer~\cite{tang2023contrastive}. This interaction occurs across all three layers in a sequential manner, resulting in progressively refined mask outputs.
\subsubsection{Dual Query Tokens}
Inspired by the application of query tokens in object detection and semantic segmentation~\cite{zhu2020deformable,tang2023contrastive}, we designed two learnable query tokens with different interaction mechanisms: the Anomaly Query Token $A$ and the Background Query Token $B$. 
In our design, the Anomaly Query Token $A$ is responsible for identifying the abnormal regions corresponding to the text, while the Background Query Token $B$ is responsible for identifying background regions unrelated to the abnormal regions indicated by the text. Initially, these two tokens are randomly initialized, and both tokens have dimensions of \(\mathbb{R}^{N \times C_t}\), where \(C_t\) is the channel dimension of the tokens. 

\begin{table*}[ht]
\vspace{-15px}
\centering
\caption{Comparisons with the state-of-the-art RIS models on our Mvrec-Ref dataset.}
\label{table1:sota}
\resizebox{\textwidth}{!}{%
\begin{tabular}{l|c|cccc|cccc}
\hline
  &  & \multicolumn{4}{c}{Val} & \multicolumn{4}{c}{Test} \\
\cline{4-10}
\multirow{-2}{*}{Model}  & \multirow{-2}{*}{Visual} & mIoU  & gIoU  & T-acc & N-acc & mIoU   & gIoU  & T-acc & N-acc \\
\hline
CRIS~\cite{wang2022cris}& CLIP-R101   & 47.34 & 61.48 & 98.41   & 90.23 & 50.04  & 66.78 & 98.32   & 94.58   \\
BKINet~\cite{ding2023bilateral}  & CLIP-R101  & 47.43 & 54.16 & 86.34   & 99.52  & 48.84  & 58.91 & 87.10   & 99.23  \\
ETRIS~\cite{xu2023bridging}   & CLIP-ViTB   & 38.50  & 54.59 & 95.46 & 98.24   & 44.55  & 63.10  & 94.12   & 98.15   \\
ADSA~\cite{yue2024adaptive} & CLIP-ViTB   & 58.32 & 69.09 & 98.35 & 99.57   & 56.88  & 65.07 & 97.67   & 97.56   \\
\hline
LAVT~\cite{yang2022lavt}   & Swin-B    & 57.93 & 68.70  & 98.13   & 98.28   & 59.72  & 72.71 & 97.45   & 96.38   \\
ReLA~\cite{liu2023gres} & Swin-B    &  63.04 & 69.86  &   98.45  &   98.56  & 60.33	   &  72.67 & 97.36  &  98.42  \\
CGFormer~\cite{tang2023contrastive} & Swin-B    & 64.51 & 70.31 & 99.48   & 97.09   & 63.40  & 73.49 & 99.35   & 97.46 \\
\hline
\textbf{DQFormer (Ours)}&  \textbf{Swin-B }   & \textbf{67.59} & \textbf{73.73} & \textbf{99.56}   & \textbf{98.67}   & \textbf{67.90}   & \textbf{77.65} & \textbf{99.38}   & \textbf{98.21}  \\
\hline
\end{tabular}
}

\vspace{-15px}
\end{table*}
The architecture of the \textbf{Mask Group Transformer Layer} is depicted in the bottom right of Fig.\ref{fig: main}. It consists of a Mask Self-Attention Layer and a Group block. 
%
\\
\noindent\textbf{Masked Self-Attention Layer} preloads linguistic information for Anomaly Query Token $A$ using a modified self-attention mechanism~\cite{vaswani2017attention}. Unlike previous methods that either directly initialize queries with language features or allow all queries to interact with language features~\cite{tang2023contrastive,yang2022lavt}, our method first maps the language feature \( F_l \) to the same dimension as the Anomaly Query Token \( A \). The mapped language feature is denoted as \( F'_l \in \mathbb{R}^{L \times C_t} \).
Next, the Anomaly Query Token \( A^{k-1} \), Background Query Token \( B^{k-1} \), and the mapped language feature \( F'_l \) are concatenated into a combined token sequence: $Q = [A^{k-1}; B^{k-1}; F'_l] \in \mathbb{R}^{(2 +  N) \times C_t},$
where $k$ means $k$-th layer. As shown in Fig.\,\ref{fig: main}, we employ a specially designed attention mask \( M' \) to ensure that only the Anomaly Query Token \( A \) interacts with the language feature \( F'_l \). This mask facilitates the interaction between \( A^{k-1} \) and \( F'_l \) while simultaneously preventing the Background Query Token \( B \) from interacting with \( F'_l \). The self-attention mechanism is then applied to the concatenated sequence \( Q \) using the attention mask \( M' \). The process involves the following steps:
\begin{equation}
\begin{aligned}
    Q_q, Q_k, Q_v &= Q W_q, \; Q W_k, \; Q W_v, \\
    Q’ &= \text{softmax}\left(\frac{Q_q Q_k^\top}{\sqrt{C_t}} + M'\right) Q_v W_o, 
\end{aligned}
\end{equation}
where \( W_q, W_k, W_v, W_o \in \mathbb{R}^{C_t \times C_t} \) are the learnable projection matrices for the query, key, value, and output, respectively. \( A^k \) and \( B^k \) correspond to the first and second dimensions of \( Q' \), respectively, matching their positions in the concatenated sequence \( Q \).
%
\\
\noindent\textbf{Group Block} facilitates cross-modal interaction by grouping aggregated visual features \( F^k_{a} \)  with the concatenated query tokens \( T^k = [A^k; B^k] \). The aggregated visual features \( F^k_a \in \mathbb{R}^{H_k \times W_k \times C_k}\) are projected into a common feature space with \( T^k \in \mathbb{R}^{2 \times  C_t} \) using learnable projection matrices:
\begin{equation}
\begin{aligned}
    \tilde{T}^k &= T^k W_t, \; \tilde{F}^k_a = \text{flatten}(F^k_a) W_d.
\end{aligned}
\end{equation}
We calculate the similarities \( S_{\text{pixel}} \in \mathbb{R}^{2 \times H_i W_i} \) between $\tilde{T}^k$ and $\tilde{F}^k_a$: $S_{\text{pixel}} = \text{norm}(\tilde{T}^k) \text{norm}(\tilde{F}^k_a)^\top.$
Next, based on the similarity \( S_{\text{pixel}} \), we group the features in \( \tilde{F}^k_a \) and  corresponds the groups to tokens \( \tilde{T}^k \) via Gumbel-softmax, thereby generating the mask \( S_{\text{mask}} \):
\begin{equation}
\begin{aligned}
    S_{\text{gumbel}} &= \text{softmax}\left(\frac{S_{\text{pixel}} + G}{\tau}\right), \\
    S_{\text{mask}} &= \text{onehot}(\text{argmax}(S_{\text{gumbel}}))^\top \\
    &\quad - \text{sg}(S_{\text{gumbel}}) + S_{\text{gumbel}},
\end{aligned}
\end{equation}
Here, \( \tau \) is a learnable coefficient, \( G \in \mathbb{R}^{2 \times H_i W_i} \) is sampled from the Gumbel(0, 1) distribution~\cite{jang2016categorical}, which allows the grouping operation to be differentiable. The term $\text{sg}$ refers to the stop gradient operator, \text{argmax} selects the corresponding token of \( \tilde{T}^k\) with the highest similarity for each feature in \( S_{\text{gumbel}} \), and the onehot operation transforms token indexes into one-hot vectors.
\( \tilde{F}^k_a \) is then integrated to update the tokens \( T^{k+1} \) using the mask \( S_{\text{mask}} \):
\begin{equation}
    T^{k+1} = \text{MLP}(S_{\text{mask}} \tilde{F}^k_a) + \tilde{T}^k.
\end{equation}

\subsubsection{Output and Loss.} Finally, the updated Anomaly Query Token \( A \) and the highest resolution aggregated visual feature map \( F^3_a \) generate the final mask output \( M \), while the Background Query Token \( B \) is mapped to predict a No-Anomaly Identifier \( E \). For intermediate layers, we use the attention map $S_{\text{mask}}$ as intermediate mask constraint to help the Dual Query Token better distinguish specified anomalies from other regions. The predicted mask $M$ and intermediate mask $S_{\text{mask}}$ are supervised by the ground-truth target mask $M_{GT}$.  Meanwhile, the No-Anomaly Identifier \( E \) is supervised by the ground-truth target label $E_{GT}$. The mask $M$ and intermediate mask $S_{\text{mask}}$ are guided by a combination of dice loss and focal loss, while $E$ is guided by the cross-entropy loss. During inference, if $E$ is predicted to be positive, the output mask $M$ will be set to empty.

\section{Experiments}
\subsection{Implementation Details}
\subsubsection{Dataset and Evaluation Metrics}
The MVTec-Ref dataset consists of $2,110$ image-expression-label triplets, with $1,224$ triplets for training, $384$ for validation, and $502$ for testing. 
Firstly, we use N-acc. and T-acc., as defined in GRES~\cite{liu2023gres}, to evaluate the model's ability to distinguish normal samples from anomalous ones.
We report our segmentation performance using three kinds of metrics: overall intersection-over-union (gIoU), generalized IoU~(gIoU), and precision at threshold values from 0.5 to 0.9. 
For brevity, detailed definitions and descriptions of the indicators are included in the \textit{Appendix}.
%
%
%

\subsubsection{Baselines} 
At present, there are no specific referring image segmentation approaches tailored for anomaly images. 
To address this gap, we evaluate seven existing methods that were originally developed for natural images. 
%
%
We conduct our experiments using the optimal models from the original papers. Additionally, we incorporate a No-Anomaly Identifier into each method to assess their ability to distinguish between normal and anomalous samples.
%
\begin{figure*}[] 
\vspace{-10px}
\centering 
\includegraphics[width=\textwidth]{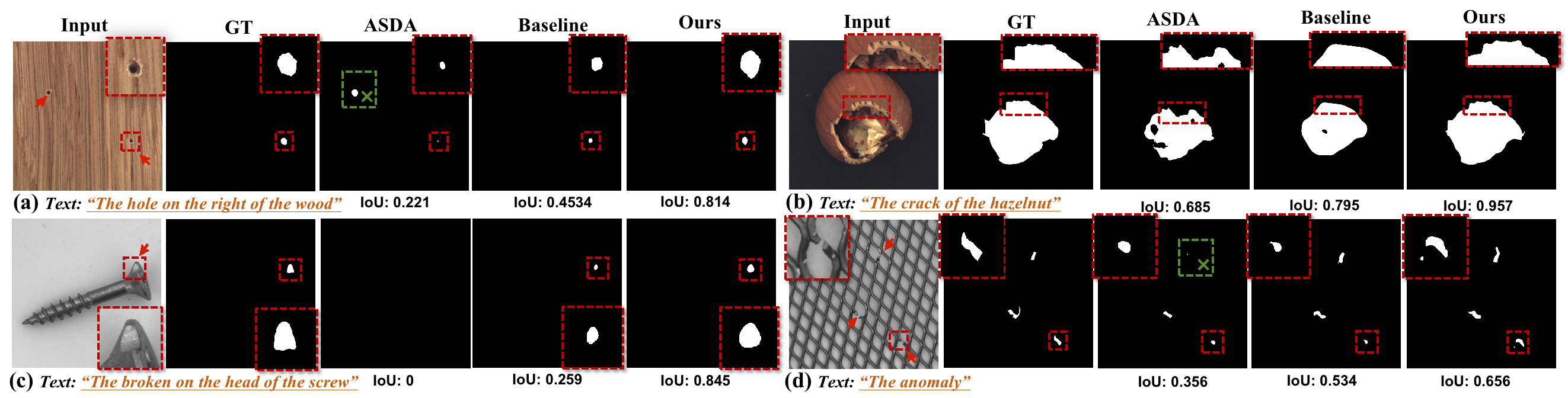} 
\vspace{-15px}
\caption{\normalfont Illustration of some segmentation results and their corresponding IoU scores from different methods.
}
\vspace{-15px}
\label{fig: Visualization} 
\end{figure*}

\subsection{Results on Mvtec-Ref}
To demonstrate the effectiveness of our DQFormer, we compare it against seven SOTA referring image segmentation methods.
As shown in Tab.\,\ref{table1:sota}, our experiments reveal that CLIP-based methods~\cite{radford2021learning} perform significantly worse on our proposed Mvtec-Ref dataset compared to the commonly used Swin backbone~\cite{liu2021swin}. 
This can be attributed to CLIP's coarse-grained approach, which is less effective at detecting the small anomalies prevalent in our dataset, where $95\%$ of images feature masks covering less than $10\%$ of the area.
%
Our DQFormer surpasses all compared methods, including CGFormer~\cite{tang2023contrastive}, with improvements of 3.08\% in mIoU and 3.42\% in gIoU on the val set, and 2.98\% in mIoU and 3.16\% in gIoU on the test set. 

To further illustrate the superiority of our DQFormer, we performed a visual analysis with corresponding IoU scores for different methods as shown in Fig.\,\ref{fig: Visualization}.
For instance, in scenario (a), our DQFormer accurately identified and localized anomalies based on text prompts, outperforming ASDA which struggled with text comprehension.
%

In the setting of unsupervised training with a large number of normal samples, given that our DQFormer trains a single model on samples from all classes within a dataset, we selected UniAD~\cite{zhao2023omnial}, which is trained under the same setup, as a baseline for comparison.
Additionally, we compare our DQFormer with DREAM~\cite{zavrtanik2021draem} and JNLD~\cite{zhao2022just} using the same unified setting.
Following existing IAD methods~\cite{chalapathy2018anomaly}, we employ the Area Under the Receiver Operating Characteristic (AUC) as our evaluation metric, with image-level and pixel-level AUC used to assess anomaly detection and anomaly localization performance, respectively.
The results on our test split are presented in Tab.\,\ref{table3:IAD}. Specific implementation details and metrics are provided in the \textit{Appendix}.

\begin{table}[]
\centering
\caption{Ablation study on MVTec-Ref test splitt.}
\label{table1:ablation}
\resizebox{0.8\textwidth}{!}{
\begin{tabular}{c|c|ccccc}
\hline
& Method                  & Pr@50 & Pr@70 & Pr@90 & mIoU  & gIoU  \\
\hline
(a) & baseline                & 78.06 & 52.23 & 15.98 & 63.40 & 73.49 \\
(b) & (a)+mask self-attention & 80.60  & 57.39 & 17.01 & 65.04 & 75.11 \\
(c) & (b)+dual query token    & 80.98 & 57.48 & 17.81 & 65.87 & 75.29 \\
(d) & (c)+local convolution   & 81.80  & 56.60  & 17.97 & 66.18 & 75.84 \\
\hline
(e) & (c)+avg pooling(full)   & 82.19 & 57.47 & 16.38 & 67.90  & 77.65 \\
\hline
\end{tabular}
}
\vspace{-28px}
\end{table}
\subsection{Ablation Study}
To validate the effectiveness of our module, we carry out ablation experiments on the test split of MVTec-Ref in Tab.\,\ref{table1:ablation}. 
\\
\noindent\textbf{Effect of Mask Self-attention and Dual Query Token.} As shown in Tab. \ref{table1:ablation}, (a) we select CGFormer~\cite{tang2023contrastive} as our baseline, which uses multiple tokens to group visual features and allows all tokens to interact with the textual features. (b) We improve this interaction mechanism by distinguishing between Anomaly tokens and other tokens, allowing only the Anomaly token to interact with the textual features. Additionally, the final mask output \( M \) is generated exclusively by the Anomaly token interacting with the highest resolution aggregated visual feature \( F^3_a \). This approach resulted in a 1.64\% improvement in mIoU and a 1.62\% improvement in gIoU. 
(c) Given that anomaly images focus on detecting patterns rather than recognizing multiple objects, we hypothesize that a single Background Token is sufficient. Unlike natural images that require multiple tokens to identify various objects, anomaly images generally consist of only abnormal and normal regions. Using a single Background Token simplifies the model while effectively capturing the necessary information. Results show that our Dual Query Token outperforms redundant query methods.
\\
\begin{table}[]
\vspace{-20px}
\caption{Comparison between unsupervised anomaly detection methods on MVTec-Ref test split.}
\label{table3:IAD}
\centering
\resizebox{0.6\textwidth}{!}{
\begin{tabular}{l|ccc}
\hline
 Method & Image-AUC  & Pixel-AUC & Accuracy   \\
\hline
JNLD~\cite{zhao2022just}  & 91.30   & 88.60  & 88.64  \\
DRAEM~\cite{zavrtanik2021draem} & 89.76  & 88.24 & 85.36  \\
UniAD~\cite{zhao2023omnial} & 96.57  & \textbf{96.9} & 91.94  \\
\hline
ours  & \textbf{97.36} & 93.13 & \textbf{97.64} \\
\hline
\end{tabular}
}
\vspace{-20px}
\end{table}

\noindent\textbf{Effect of multi-level local feature aggregation in LMA.} To better model anomalies of varying sizes and shapes, we implement multi-level local feature aggregation on the visual features in LMA. We experiment with different approaches, including varying convolution kernel sizes (d) and average pooling with different receptive fields (e). Our comparisons reveal that parameter-free average pooling performs better, surpassing (d) by 1.72\% in mIoU and 1.81\% in gIoU. This improvement is likely due to the use of pretrained weights from the Swin transformer~\cite{liu2021swin}, where the self-attention mechanisms already capture global information. The average pooling operation enhances this by enabling multi-scale perception of neighboring features.


\section{Conclusion}
In this paper, we introduce Referring Industrial Anomaly Segmentation (RIAS), a novel paradigm that leverages language to guide anomaly detection. To support RIAS, we developed the MVTec-Ref dataset, specifically designed to reflect real-world industrial environments with diverse, scenario-specific referring expressions. To evaluate the effectiveness of our proposed paradigm and dataset, we present a benchmark framework called Dual Query Token with Mask Group Transformer (DQFormer), enhanced by Language-Gated Multi-Level Aggregation (LMA). We believe that RIAS, equipped with the MVTec-Ref dataset, will advance Industrial Anomaly Detection (IAD) towards an open-set framework.

\section*{Acknowledgments}
This work was supported by National Science and Technology Major Project (No. 2022ZD0118201), the National Science Fund for Distinguished Young Scholars (No.62025603), the National Natural Science Foundation of China (No. U21B2037, No. U22B2051, No. U23A20383, No. 62176222, No. 62176223, No. 62176226, No. 62072386, No. 62072387, No. 62072389, No. 62002305 and No. 62272401), and the Natural Science Foundation of Fujian Province of China (No.2022J06001).

%
%
%
\bibliographystyle{splncs04}
\bibliography{prcv}

\clearpage
\appendix

\title{Referring Industrial Anomaly Segmentation\\(\textit{Technical Appendix})}
\author{}
\institute{}


\maketitle             

\section{Table of contents}
\begin{itemize}
    \item \textcolor{red}{\S B}: Implementation Details of the Experiments
    \item \textcolor{red}{\S C}:  Additional Dataset Details
    \item \textcolor{red}{\S D}: Additional Experimental Results
    \item \textcolor{red}{\S E}: Additional Visualization Results
\end{itemize}

\section{Implementation Details of the Experiments}
\textit{DQFormer} is implemented in PyTorch and trained using the Adam optimizer. During training, the batch size is set to 16, with an initial learning rate of 1e-4. The entire training process takes approximately 100 minutes for 50 epochs on 4 RTX 3090 GPUs. We will provide our training code and experiment logs in the open-source release.

For the other RIS models used for comparison, we conduct our experiments using the optimal settings from the original papers. Additionally, we integrat a No-Anomaly Identifier into each method to evaluate their ability to distinguish between normal and anomalous samples. We design the No-Anomaly Identifier for each model based on both experience and multiple experimental trials to ensure a fair comparison.

We use N-acc. and T-acc. to evaluate the model's ability to distinguish normal samples from anomalous ones.
These metrics assess the model’s performance in no-anomaly identification. 
For a no-anomaly sample, a prediction with no anomaly pixels is a true positive (\(TP\)), otherwise, it is a false negative (\(FN\)). N-acc. measures the model’s performance in identifying no-anomaly samples:
$
\text{N-acc.} = \frac{TP}{TP + FN}
$.
T-acc. measures how generalization on no-anomaly samples impacts the performance on anomaly samples, indicating the number of samples with anomalies that are misclassified as no-anomaly:
$
\text{T-acc.} = \frac{TN}{TN + FP}
$.
We report our segmentation performance using three kinds of metrics: overall intersection-over-union (gIoU), generalized IoU~(gIoU), and precision at threshold values from 0.5 to 0.9. oIoU is calculated by the ratio of the total intersection area to the total union area across all test samples, while gIoU is the average of IoU values between predicted masks and ground truths for all test samples. For no-target samples, the IoU values of true positive no-target samples are regarded as 1, while IoU values of false negative samples are treated as 0.

\begin{figure}[t]
    \centering
    \begin{subfigure}[b]{0.45\textwidth} 
        \centering
        \includegraphics[width=\textwidth]{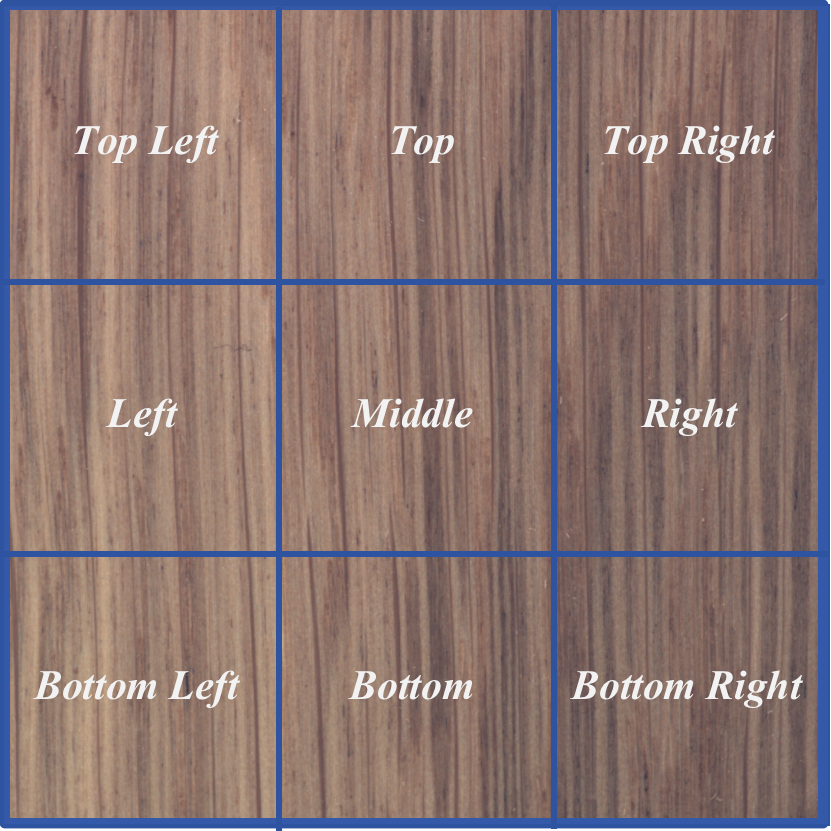}
        \caption{Definition of orientation for textured objects.}
        \label{fig:sub1}
    \end{subfigure}
    \hfill
    \begin{subfigure}[b]{0.45\textwidth} 
        \centering
        \includegraphics[width=\textwidth]{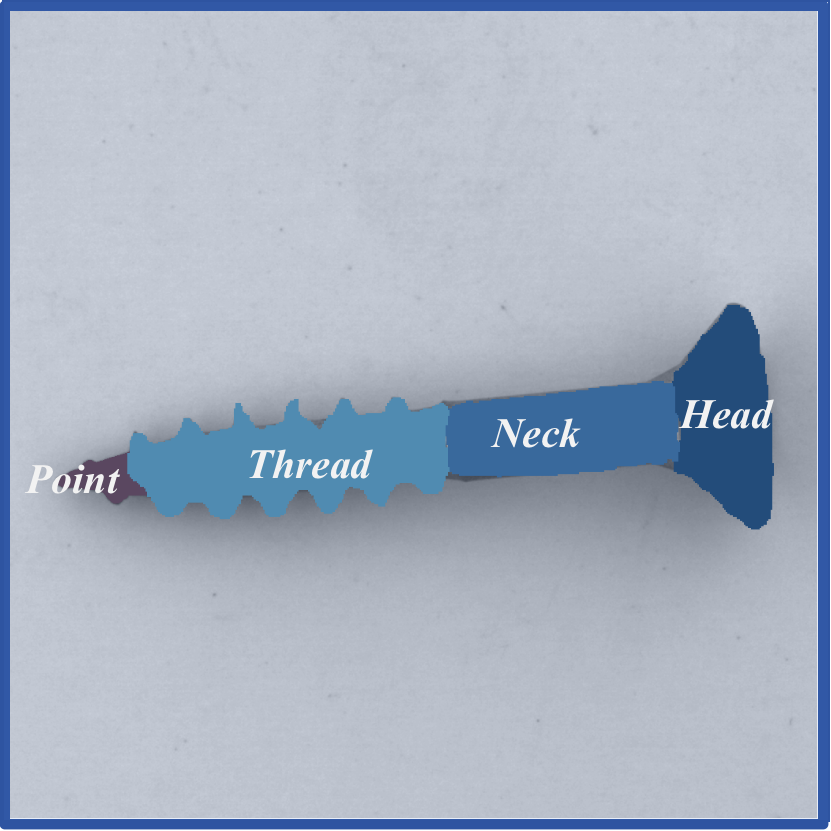}
        \caption{Definitions of  position in the screw.}
        \label{fig:sub2}
    \end{subfigure}
    \vspace{-6px}
    \caption{Definitions of (a) orientation for textured objects and (b) position in the screw.}
    \label{fig:overall}
\vspace{-15px}
\end{figure}

\begin{figure}[t]
    \centering
    \begin{subfigure}[b]{\textwidth} 
        \centering
\includegraphics[width=\textwidth]{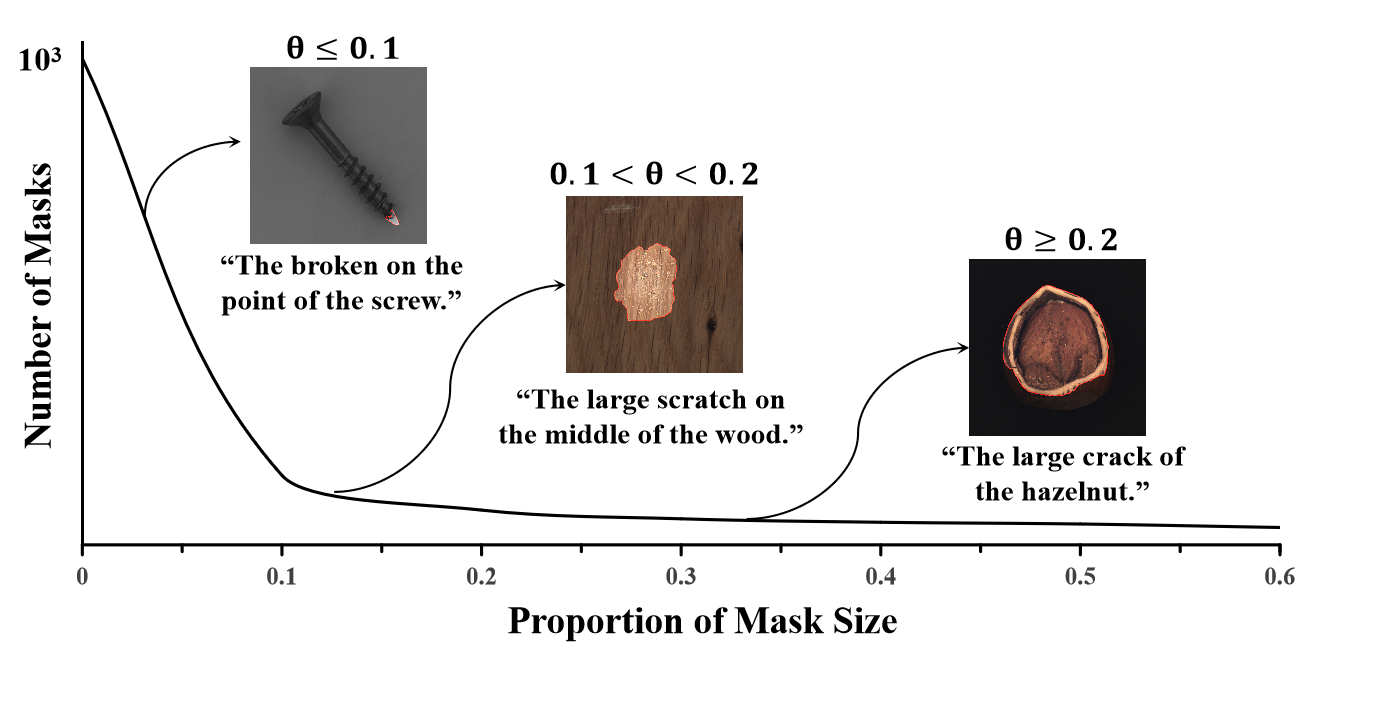}
        \vspace{-30px}
        \caption{Anomaly mask size distribution.}
        \label{fig:large}
    \end{subfigure}
    

    \begin{subfigure}[b]{0.45\textwidth} 
        \centering
    \includegraphics[width=\textwidth]{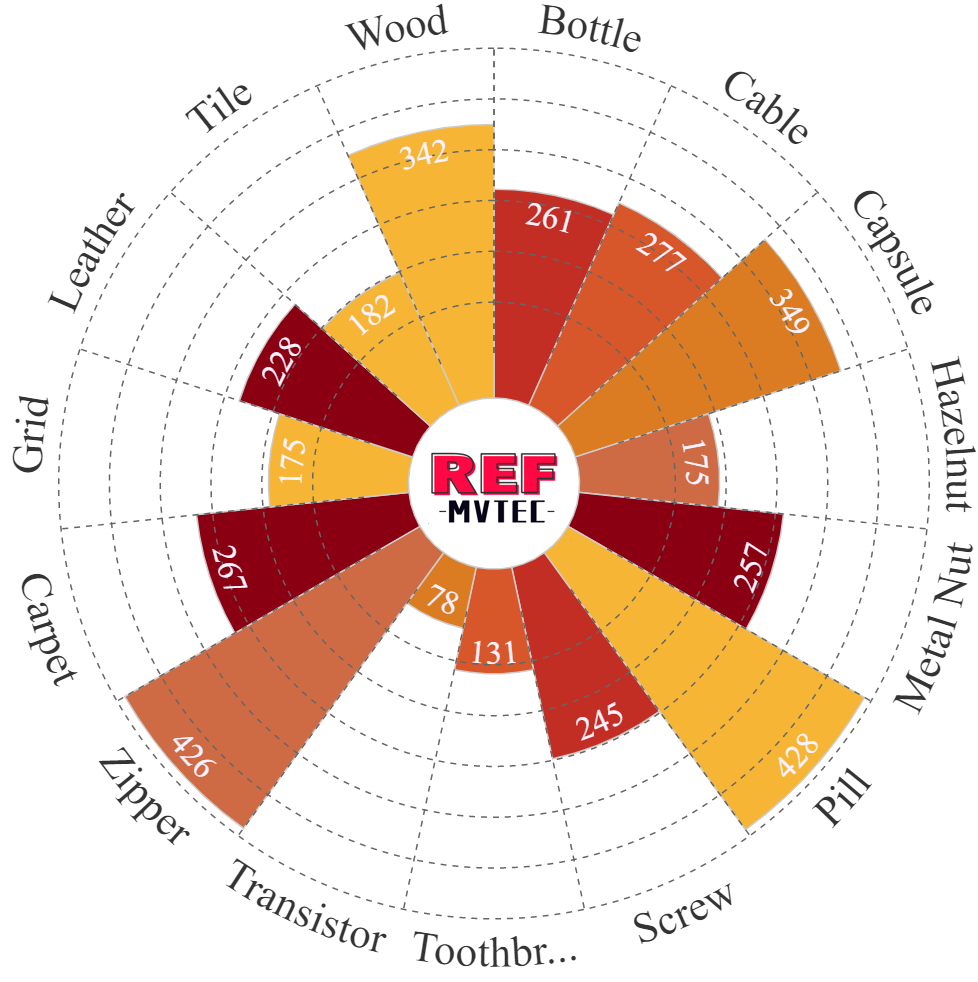}
        \caption{Object category 
 distribution.}
        \label{fig:small1}
    \end{subfigure}
    \hfill
    \begin{subfigure}[b]{0.45\textwidth} 
        \centering
        \includegraphics[width=\textwidth]{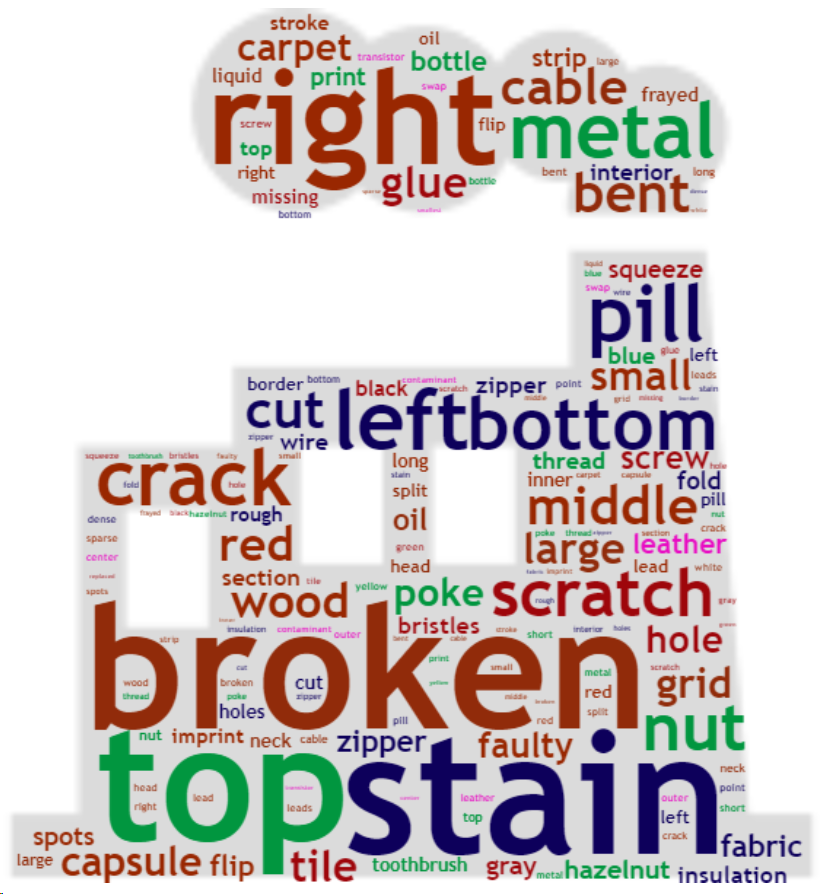}
        \caption{Word cloud.}
        \label{fig:small2}
    \end{subfigure}
    
    \caption{Analysis of the dataset.}
    \label{fig:three_subfigures}
\vspace{-30px}
\end{figure}

\section{Additional Dataset Details}

\subsection{Characteristics of Anomaly Data}
\noindent \textbf{1) Category}: Anomalies are often complex and involve multiple factors, making it difficult to classify them into broad categories. For example, in the MVTec AD dataset, it's common to find multiple types of anomalies within a single image, making the identification of mixed anomalies more challenging. As illustrated in the Fig.\,\ref{fig: dataset analysis}, the typical approach of assigning a singular `combined' label to images containing multiple anomaly types does not facilitate detailed anomaly studies. To address this, we have refined the classification strategy using natural language descriptions to achieve more precise categorization. In the case of the wood category, specific anomalies such as knots are distinctly described. This method also allows for the flexible articulation of anomaly combinations, exemplified by descriptions like ``The knot and the red stain on the wood".
\begin{figure}[]
\centering 
\includegraphics[width=0.7\linewidth]{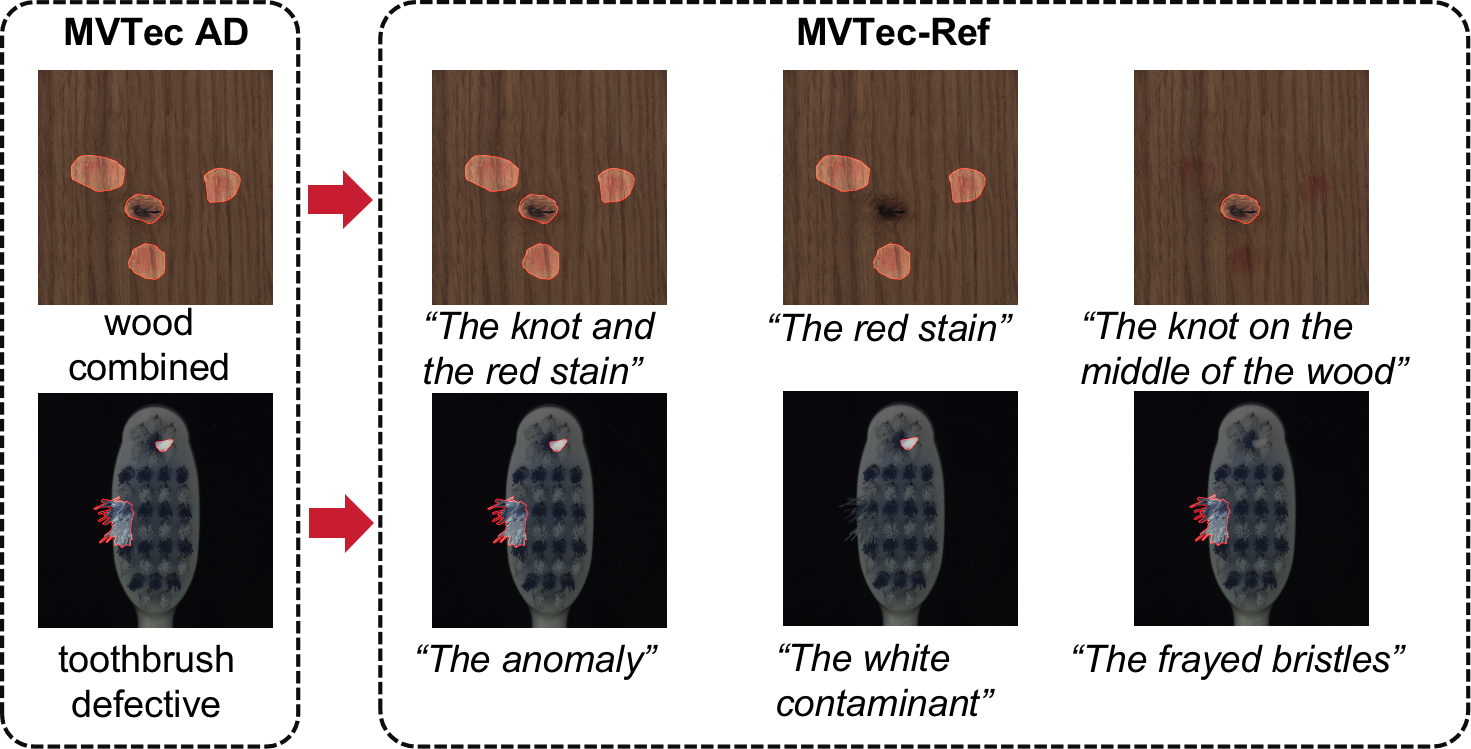} 
\hfill 
\centering
\caption{Examples from MvTec AD to MVTec-Ref.} 

\label{fig: dataset analysis} 
\end{figure}

\noindent \textbf{2) Spatial Position}: The precision of spatial position descriptions is crucial for effective anomaly detection. Traditionally, general spatial descriptors such as ``above" and ``left" were used to denote the locations of texture anomalies. However, these descriptors are often inadequate for objects subject to orientation changes, like screws. To enhance accuracy, we propose a more specific method for spatial descriptions by segmenting the object into its constituent parts. For instance, a screw can be divided into the head, neck, thread, and point. This segmentation allows for more precise descriptions of anomalies in relation to specific parts of the object, significantly improving the ease and accuracy of anomaly localization.
\begin{figure*}[ht] 
\centering 
\includegraphics[width=\textwidth]{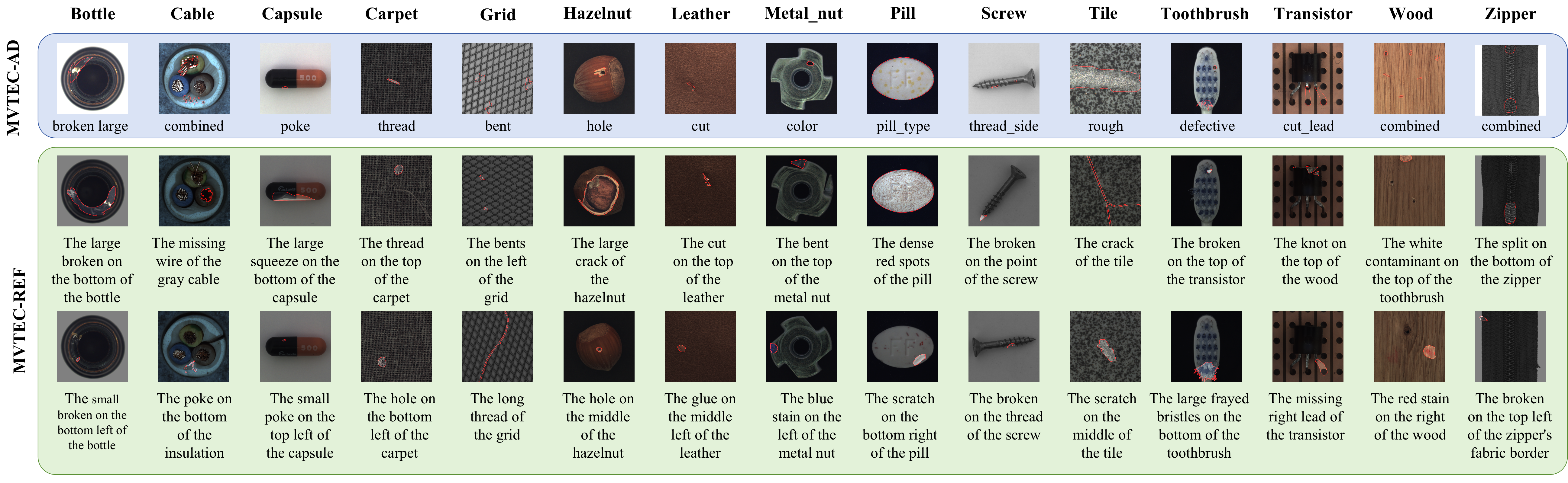} 
\caption{\normalfont Examples from our MVTec-Ref dataset. Best viewed with zoom.
}
\vspace{-10px}
\label{fig1} 
\end{figure*}

\noindent\textbf{3) Generalization}: We added general prompts like ``The anomaly'' in the annotations, which is crucial for anomaly detection tasks. This design improves the model's generalization ability, preventing it from overfitting to specific categories, which is important in general scenarios.
In general scenarios, being able to distinguish anomalies without being limited to specific categories is very important. 

Through these improvements, the MVTec-Ref dataset not only retains the strengths of MVTec-AD but also addresses its shortcomings in anomaly categorization. This enhances the dataset's applicability in complex anomaly detection tasks. The detailed and specific descriptions allow the dataset to better support the RIAS task, thereby improving the model’s effectiveness and reliability in real-world industrial inspections.

\subsection{Details of Referring Expression Annotation}

Given that end users often refer to objects by their categories, attributes, and spatial relationships with other entities, we generate linguistic expressions using the following templates. In this section, we provide more detailed annotation information compared to the paper.

\noindent \textbf{1) Category or Category with Attribute}: This expression template is commonly used when users want to specify their interest in particular objects. We generate referring expressions by either directly specifying categories like "stain" and "hole" or by adding attributes for more detail, such as "red stain." The distribution of object and anomaly categories within the text can be seen in the word cloud in Fig.\,\ref{fig:three_subfigures} (c).


\noindent\textbf{2) Category with Spatial Relation in the Object}: This template is expanded on the aforementioned spatial descriptions.  
Traditionally, general spatial descriptors such as ``above" and ``left" were used to denote the locations of texture anomalies. 
We further provide clear definitions, as shown in Fig.\,\ref{fig:overall} (a). We divided the image into nine regions, assigning spatial terms to each. During annotation, the main spatial location relative to the object's position is used. 
However, these descriptors are often inadequate for objects subject to orientation changes, like screws. 
To enhance accuracy, we propose a more specific method for spatial descriptions by segmenting the object into its constituent parts. 
As shown in Fig.\,\ref{fig:overall} (b), a screw can be divided into the head, neck, thread, and point. 
This segmentation allows for more precise descriptions of anomalies in relation to specific parts of the object, significantly improving the ease and accuracy of anomaly localization. 
As shown in \ref{fig:overall}, For example, ``The broken on the head of the screw'' combines the category with specific spatial information.
\\

\subsection{Dataset Features and Statistics}

As shown in Fig.\,\ref{fig:three_subfigures}, we conducted a thorough analysis of the dataset. (a) reveals that over 90\% of anomaly masks occupy less than 10\% of the image area, making the detection of small objects particularly challenging. (b) highlights the significant long-tail distribution of object categories, which poses a challenge for developing a unified model that comprehends all categories effectively. Lastly, (d) presents a word cloud analysis, showcasing the diversity of our textual annotations. 


\section{Additional Experimental Results}
In Tab.\,\ref{table: sup}, we present the detailed results of our DGFormer on each category in the MVTec-Ref test split. Performance varies across categories due to the differing types and difficulties of anomalies. For example, hazelnut and metal nut show very high mIoU and oIoU scores, indicating effective segmentation of prominent and easily describable anomalies, as seen in Fig.\,\ref{fig1}. 
\begin{table*}[ht]
\vspace{-20px}
\caption{Detailed results of our DGFormer on each category in the MVTec-Ref test split.}
\label{table: sup}
\centering
\resizebox{0.9\textwidth}{!}{%
\begin{tabular}{c|ccccccc|cc}
\hline
Class      & {Pr@50} & {Pr@60} & {Pr@70} & {Pr@80} & {Pr@90} & {mIoU}  & {oIoU}  & {T-acc} & {N-acc} \\
\hline
bottle     & 97.44 & 91.03 & 88.46 & 38.46 & 5.13  & 77.20  & 79.19 & 100.00   & 100.00   \\
cable      & 81.90  & 71.43 & 58.10  & 40.00 & 26.67 & 67.33 & 64.13 & 97.83 & 100.00   \\
capsule    & 76.92 & 62.39 & 37.61 & 11.11 & 5.13  & 60.09 & 54.89 & 98.20  & 100.00   \\
carpet     & 82.98 & 69.15 & 53.19 & 29.79 & 11.70  & 68.95 & 72.28 & 100.00   & 100.00   \\
grid       & 62.12 & 40.91 & 21.21 & 9.09  & 7.58  & 57.47 & 55.75 & 100.00   & 100.00   \\
hazelnut   & 100.00   & 93.24 & 85.14 & 71.62 & 35.14 & 84.90  & 89.14 & 100.00   & 100.00   \\
leather    & 79.12 & 60.44 & 39.56 & 21.98 & 7.69  & 66.03 & 66.91 & 97.62 & 100.00   \\
metal nut     & 97.80  & 91.21 & 81.32 & 62.64 & 26.37 & 81.25 & 93.21 & 100.00   & 100.00   \\
pill       & 90.44 & 83.82 & 69.85 & 48.53 & 16.18 & 75.65 & 92.95 & 99.23 & 100.00   \\
screw      & 60.61 & 48.48 & 31.31 & 16.16 & 9.09  & 52.97 & 52.37 & 97.56 & 88.89 \\
tile       & 96.00 & 85.33 & 84.00 & 84.00 & 53.33 & 85.39 & 88.02 & 100.00   & 100.00   \\
toothbrush & 70.00 & 23.33 & 16.67 & 10.00 & 10.00 & 56.60  & 60.30  & 100.00   & 100.00   \\
transistor & 77.78 & 72.22 & 66.67 & 52.78 & 44.44 & 72.62 & 72.29 & 93.67 & 100.00   \\
wood       & 79.21 & 70.30  & 53.47 & 30.69 & 16.83 & 66.98 & 72.03 & 100.00   & 100.00   \\
zipper     & 91.16 & 75.51 & 52.38 & 25.85 & 4.76  & 68.71 & 68.88 & 100.00   & 100.00  \\
\hline
all        & 82.19 & 71.62 & 57.47 & 38.71 & 16.38 & 67.90  & 77.65 & 99.38 & 98.21 \\
\hline
\end{tabular}
}
\vspace{-10px}
\end{table*}

However, DGFormer struggles with capsule and screw categories, where anomalies are often small and difficult to detect. This is reflected in the lower scores and the challenging examples shown in Fig.\,\ref{fig1}. Additionally, while many categories achieve perfect T-acc and N-acc, the N-acc for screws is only 88\%, suggesting that DGFormer tends to misclassify minor imperfections in screws as anomalies.


\end{document}